\documentclass[sigconf]{acmart}

\usepackage[ruled,noline]{algorithm2e}
\let\mc\multicolumn

\AtBeginDocument{%
  \providecommand\BibTeX{{%
    \normalfont B\kern-0.5em{\scshape i\kern-0.25em b}\kern-0.8em\TeX}}}

\copyrightyear{2021}
\acmYear{2021}
\setcopyright{iw3c2w3}
\acmConference[WWW '21]{Proceedings of the Web Conference 2021}{April 19--23, 2021}{Ljubljana, Slovenia}
\acmBooktitle{Proceedings of the Web Conference 2021 (WWW '21), April 19--23, 2021, Ljubljana, Slovenia}
\acmPrice{}
\acmDOI{10.1145/3442381.3450141}
\acmISBN{978-1-4503-8312-7/21/04}

\begin{document}

\title{Inductive Entity Representations from Text via Link Prediction}

\author{Daniel Daza}
\orcid{0000-0002-5357-3705}
\affiliation{%
  \institution{Vrije Universiteit Amsterdam}
  \institution{University of Amsterdam}
  \institution{Discovery Lab, Elsevier}
  \city{Amsterdam}
  \country{The Netherlands}
}
\email{d.dazacruz@vu.nl}

\author{Michael Cochez}
\orcid{0000-0001-5726-4638}
\affiliation{%
  \institution{Vrije Universiteit Amsterdam}
  \institution{Discovery Lab, Elsevier}
  \city{Amsterdam}
  \country{The Netherlands}
}
\email{m.cochez@vu.nl}

\author{Paul Groth}
\orcid{0000-0003-0183-6910}
\affiliation{%
  \institution{University of Amsterdam}
  \institution{Discovery Lab, Elsevier}
  \city{Amsterdam}
  \country{The Netherlands}
}
\email{p.groth@uva.nl}


\begin{abstract}
Knowledge Graphs (KG) are of vital importance for multiple applications on the web, including information retrieval, recommender systems, and metadata annotation. 

Regardless of whether they are built manually by domain experts or with automatic pipelines, KGs are often incomplete. To address this problem, there is a large amount of work that proposes using machine learning to complete these graphs by predicting new links. Recent work has begun to explore the use of textual descriptions available in knowledge graphs to learn vector representations of entities in order to preform link prediction. However, the extent to which these representations learned for link prediction generalize to other tasks is unclear. This is important given the cost of learning such representations. Ideally, we would prefer representations that do not need to be trained again when transferring to a different task, while retaining reasonable performance.

Therefore, in this work, we propose a holistic evaluation protocol for entity representations learned via a link prediction objective. We consider the \emph{inductive} link prediction and entity classification tasks, which involve entities \emph{not} seen during training. We also consider an information retrieval task for entity-oriented search. We evaluate an architecture based on a pretrained language model, that exhibits strong generalization to entities not observed during training, and outperforms related state-of-the-art methods (22\% MRR improvement in link prediction on average). We further provide evidence that the learned representations transfer well to other tasks without fine-tuning. In the entity classification task we obtain an average improvement of 16\% in accuracy compared with baselines that also employ pre-trained models. In the information retrieval task, we obtain significant improvements of up to 8.8\% in NDCG@10 for natural language queries. We thus show that the learned representations are not limited KG-specific tasks, and have greater generalization properties than evaluated in previous work.
\end{abstract}



\maketitle

\section{Introduction}

Knowledge graphs provide a structured way to represent information in the form of entities and relations between them \citep{fensel2020knowledge}. They have become central to a variety of tasks in the Web, including information retrieval \citep{dalton2014entity,gerritse2020graph}, question answering \cite{vakulenko2019message,huang2019knowledge}, and information extraction \citep{mintz2009distant,bosselut2019comet,gupta2019care}. Many of these tasks can benefit from distributed representations of entities and relations, also known as embeddings.

A large body of work in representation learning in KGs \citep{nickel2015review,wang2017knowledge} is based on the optimization of a link prediction objective, which results in embeddings that model relations in a vector space. These approaches are often touted as an alternative to logic-based systems for inference in incomplete KGs, as they can assign a score to missing links \citep{drumond2012predict,hamilton2018queries} They have also been proposed for implementing approximate forms of reasoning for the Semantic Web~\cite{hitzler2020neural}. However, by design, some of these methods can only compute predictions involving entities observed during training. This results in approaches that fail when applied to real-world, dynamic graphs where new entities continue to be added.

\begin{figure}[t]
\centering
\includegraphics[width=0.90\linewidth]{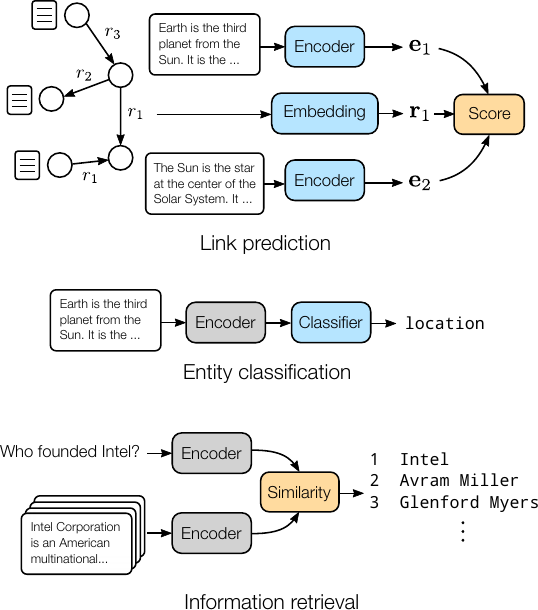}
\caption{Overview of our work: using entity descriptions, an entity encoder is trained for link prediction in a knowledge graph (top). We show that the encoder can then be used without fine-tuning to obtain informative features for entity classification (middle) and information retrieval (bottom).}
\label{fig:model}
\end{figure}

To overcome this challenge, we look to make use of the textual information within KGs. KGs like YAGO \citep{suchanek2007yago}, DBpedia \citep{auer2007dbpedia}, and industry-deployed KGs \citep{dong2014knowledge,noy2019industry}, contain rich textual attributes about entities such as names, dates, and descriptions \citep{fensel2020knowledge}.  Therefore, it seems reasonable to assume that for real-world applications, attribute data such as entity descriptions are readily available.

From this perspective, methods that treat KGs merely as a collection of nodes and labeled links are needlessly discarding a valuable source of information. Previous work has proposed to use textual descriptions for learning entity representations \cite{shi2018open,yao2019kg,xie2016RepresentationLO,wang2019kepler}, which results in a much more flexible approach, since entity representations are computed as a \emph{function} of their textual description, and can thus be obtained even for entities not observed during training. Unfortunately, the evaluation protocol in these works mainly focuses on the task of link prediction, leaving other potential outcomes of such a flexible approach unexplored.

The motivation for seeking representations that generalize well is that they can be applied in a variety of settings for which they were not explicitly trained, while retaining reasonable performance. This avoids having to invest more resources on data collection, labelling, and fine-tuning when faced with a new task.

In this work, we are thus interested in the following research question: \textbf{What are the generalization properties of entity representations learned via a link prediction objective?}
Our work towards answering this question results in the following contributions:

\begin{enumerate}
\item We propose the use of a pretrained language model for learning representations of entities via a link prediction objective, and study its performance in combination with four different relational models.
\item We propose a holistic evaluation framework for entity representations, that comprises link prediction, entity classification, and information retrieval.
\item We provide evidence that entity representations based on pretrained language models exhibit strong generalization properties across all tasks, outperforming the state-of-the-art, and are thus not limited to KG-specific tasks.
\end{enumerate}

The rest of this paper is organized as follows. In \autoref{sec:related} we discuss the related work. In \autoref{sec:inductive} we introduce the need for learning inductive entity representations, and motivate the use of a pretrained language model for the task. In \autoref{sec:experiments} we describe the experiments and results for the three tasks mentioned above. Finally, we conclude and highlight directions of future work.  

\section{Related work}
\label{sec:related}
Multiple methods in the literature of representation learning in KGs propose to learn an embedding for each entity and relation in a KG. Well known examples include RESCAL~\citep{nickel2011three}, TransE \citep{bordes2013translating}, and DistMult \citep{yang2015embedding}, among others~\citep{nickel2015review,wang2017knowledge}. While the state of the art in the task of link prediction continues to improve \citep{ruffinelli2020you}, most models essentially learn a lookup table of embeddings for entities and relations, and thus they are not applicable to the scenario where new entities are added to the graph.

A natural way to avoid this problem is to train \textit{entity encoders}, that operate on a vector of entity attributes. Such encoders have been implemented using feed-forward and graph neural networks \citep{cao2016deep,hamilton2017inductive,kipf2016variational,schlichtkrull2018modeling}. While they can produce representations for new entities, they require fixing a set of attributes before training (e.g. bag-of-words, or numeric properties) restricting the domain of application. Furthermore, as recently proposed inductive methods~\cite{teru2019inductive}, they can only produce representations for new entities, \emph{using their relations with existing entities}, which is unsuitable for inductive link prediction, particularly in the challenging setting where all entities were not seen during training.

Recent work has explored using textual descriptions of entities and relations for link prediction, and proposes architectures to assign a score given the description of a relation and the entities involved in it \citep{shi2018open,yao2019kg}. However, these architectures take as input simultaneously descriptions of entities and relations and output a score. This unavoidably mixes entity and relation representations, and prevents their transfer to other tasks such as entity classification and information retrieval.

The closest methods to our work are based on the idea of training an entity encoder with a link prediction objective. DKRL~\cite{xie2016RepresentationLO} consists of a convolutional neural network (CNN) that encodes descriptions. The performance of this method is limited as it does not take stop words into account, which discards part of the semantics in entity descriptions. Furthermore, its CNN architecture lags behind recent developments in neural networks for natural language processing, such as self-attention~\cite{vaswani2017attention}.

Pre-trained language models that use self-attention, such as BERT \citep{devlin2019bert} have be shown to be effective at capturing similarity between texts using distributed representations~\citep{reimers2019sentence,petroni2019zero}. In KEPLER~\cite{wang2019kepler}, the authors propose a model that uses BERT as an entity encoder, trained with an objective that combines language modeling and link prediction. The language modeling objective translates into increased training times, computing resources, and the requirement of a large corpus with long entity descriptions, of up to 512 tokens. In our work, we propose to use a pre-trained language model trained exclusively for link prediction, and obtain significant improvements at a reduced computational cost.

The evaluation protocol for both DKRL and KEPLER contains two fundamental issues that we address here. Firstly, the methods are implemented with a translational relational model~\cite{bordes2013translating}. However, in principle this is not necessarily the best model for any description encoder, and so it remains an open question if other models, such as multiplicative interaction models~\citep{yang2015embedding,trouillon2016complex}, are better suited. We address this by considering four different relational models in our experiments, and show that a choice of model does matter.

Secondly, the evaluation of generalization in these works is limited. In DKRL, the entity representations are evaluated in a limited inductive link prediction setting, and in an entity classification task where entities in the test set were also used for training. In KEPLER, the authors only consider the link prediction task. In our work, we detail a more extensive evaluation framework that addresses these issues, including two different formulations of the inductive setting for link prediction.

\section{Inductive entity representations}\label{sec:inductive}

We define a \textit{knowledge graph with entity descriptions} as a tuple $\mathcal{G} = (\mathcal{E}, \mathcal{R}, \mathcal{T}, \mathcal{D})$, consisting of a set of entities $\mathcal{E}$, relation types $\mathcal{R}$, triples $\mathcal{T}$, and entity descriptions $\mathcal{D}$. Each triple in $\mathcal{T}$ has the form $(e_i, r_j, e_k)$, where $e_i\in\mathcal{E}$ is the head entity of the triple, $e_k\in\mathcal{E}$ the tail entity, and $r_j\in\mathcal{R}$ the relation type. For each entity $e_i\in\mathcal{E}$, there exists a description $d_{ei} = (w_1, \dots,w_n)\in\mathcal{D}$, where all $w_i$ are words in a vocabulary $\mathcal{V}$.

For an entity $e_i\in\mathcal{E}$, we denote its embedding as a vector $\mathbf{e}_i\in\mathbb{R}^d$, and similarly $\mathbf{r}_j\in\mathbb{R}^d$ for the embedding of a relation $r_j\in\mathcal{R}$, where $d$ is the dimension of the embedding space. We consider the problem of optimizing the embeddings of entities and relations in the graph via link prediction, so that a scoring function $s(\mathbf{e}_i, \mathbf{r}_j, \mathbf{e}_k)$ assigns a high score to all observed triples $(e_i, r_j, e_k)\in\mathcal{T}$, and a low score to triples not in $\mathcal{T}$. This can be achieved by minimizing a loss function such as a margin-based loss \citep{bordes2011learning,bordes2013translating},
\begin{equation}
    \sum_{(e_i,r_j,e_k)\in\mathcal{T}}\max(0, 1 - s(\mathbf{e}_i, \mathbf{r}_j, \mathbf{e}_k) + s(\mathbf{e}_i', \mathbf{r}_j, \mathbf{e}_k')),
    \label{eq:margin-loss}
\end{equation}
where $\mathbf{e}_i'$ and $\mathbf{e}_k'$ are embeddings for an unobserved \textit{negative triple} $(e_i', r_j, e_k')\notin\mathcal{T}$. Other suitable loss functions include the binary and multi-class cross-entropy \citep{trouillon2016complex,kadlec2017strike}.

In general, for each triple in the KG, these loss functions can be written in the form $\mathcal{L}(s_p, s_n)$, as a function of the score $s_p$ for a positive triple, and $s_n$ for a negative triple. We list some of the scoring functions proposed in the literature in Table \ref{tab:scoring}.

\begin{table}
\caption{Examples of scoring functions for triples, proposed for TransE \citep{bordes2013translating}, DistMult \citep{yang2015embedding}, ComplEx \citep{trouillon2016complex}, and SimplE \citep{kazemi2018simple}. For a triple $(e_i, r_j, e_k)$, we denote as $\mathbf{e}_i$, $\mathbf{r}_j$ and $\mathbf{e}_k$ the embeddings of its constituents (in SimplE these have two parts that we indicate with indices). $\Vert\cdot\Vert_p$ indicates the $p$-norm; $\langle\cdot, \cdot, \cdot \rangle$ is the generalized three-way dot product; $\operatorname{Re}(\cdot)$ is the real part of a complex number; and $\bar{\mathbf{e}_k}$ is the complex conjugate of a complex-valued vector $\mathbf{e}_k$.}
\label{tab:scoring}
\centering
\begin{tabular}{lc}
\toprule
Model                                & Function \\
\midrule
TransE   & $-\Vert \mathbf{e}_i + \mathbf{r}_j - \mathbf{e}_k  \Vert_p$                      \\
DistMult & $\langle\mathbf{e}_i, \mathbf{r}_j, \mathbf{e}_k\rangle$                          \\
ComplEx  & $\operatorname{Re}(\langle\mathbf{e}_i, \mathbf{r}_j, \bar{\mathbf{e}_k}\rangle)$ \\
SimplE   & $\frac{1}{2}\left(\langle\mathbf{e}_{i1}, \mathbf{r}_{j1}, \mathbf{e}_{k1}\rangle + \langle\mathbf{e}_{i2}, \mathbf{r}_{j2}, \mathbf{e}_{k2}\rangle\right)$ \\
\bottomrule
\end{tabular}
\end{table}

The previous optimization objective is pervasive in \textit{transductive} methods for representation learning in knowledge graphs \citep{nickel2015review,wang2017knowledge,ruffinelli2020you}, which are limited to learning representations for entities in a fixed set $\mathcal{E}$. In these methods, the entity and relation embeddings are optimized when iterating through the set of \textit{observed} triples. Therefore, by design, {\em prediction at test time is impossible for entities not seen during training}.

We can circumvent this limitation by leveraging statistical regularities present in the description of entities~\citep{xie2016RepresentationLO,shi2018open,wang2019tackling,wang2019kepler}. This can be realized by specifying a parametric \textit{entity encoder} $f_\theta$ that maps the description $d_{ei}$ of an entity to a vector $\mathbf{e}_i = f_\theta(d_{ei})\in\mathbb{R}^d$ that acts as the embedding of the entity. The learning algorithm is then carried out as usual, by optimizing the parameters $\theta$ of the entity encoder and the relation embeddings $\mathbf{r}_j\ \forall r_j \in \mathcal{R}$, with a particular score and loss function. This process allows the encoder to learn \textit{inductive entity representations}, as it can embed entities not seen during training, as long as they have an associated description.

\subsection{BERT for entity descriptions}
\label{sec:blpdef}

Transformer networks \citep{vaswani2017attention} have been shown to be powerful encoders that map text sequences to contextualized vectors, where each vector contains information about a word in context \citep{ethayarajh2019contextual}. Furthermore, pre-trained language models like BERT \citep{devlin2019bert}, which have been optimized with large amounts of text, allow fine-tuning the encoder for a different task that benefits from the pre-training step.

We select BERT for the entity encoder in our method, but other pre-trained models based on Transformers are equally applicable. Note that unlike DKRL~\cite{xie2016RepresentationLO}, this entity encoder is well suited for inputs in natural language, rather than processed inputs where stop words have been removed. We expect accepting raw inputs helps the encoder better capture the semantics needed to learn more informative entity representations.

Given an entity description $d_{ei} = (w_1, \dots, w_n)$, the encoder first adds special tokens \texttt{[CLS]} and \texttt{[SEP]} to the beginning and end of the description, respectively, so that the input to BERT is the sequence $\hat{d}_{ei} = (\texttt{[CLS]}, w_1, \dots, w_n, \texttt{[SEP]})$. The output is a sequence of $n + 2$ contextualized embeddings, including those corresponding to the added special tokens:
\begin{equation}
\texttt{BERT}(\hat{d_{ei}}) = (\mathbf{h}_\texttt{CLS}, \mathbf{h}_1, \dots, \mathbf{h}_n, \mathbf{h}_\texttt{SEP}).
\end{equation}

Similarly as in works that employ BERT for representations of text \citep{reimers2019sentence,wang2019kepler}, we select the contextualized vector $\mathbf{h}_\texttt{CLS}\in\mathbb{R}^h$, where $h$ is the hidden size of BERT. This vector is then passed through a linear layer that reduces the dimension of the representation, to yield the output entity embedding $\mathbf{e}_i = \mathbf{W}\mathbf{h}_{\texttt{CLS}}$, where $\mathbf{W}\in\mathbb{R}^{d\times h}$ is a parameter.

For relation embeddings, we use randomly initialized vectors $\mathbf{r}_j\in\mathbb{R}^d$, for each $r_j\in\mathcal{R}$. We then apply stochastic gradient descent to optimize the model for link prediction: for each positive triple $(e_i, r_j, e_k)\in\mathcal{T}$ we compute a positive score $s_p$. By replacing the head or the tail with a random entity, we obtain a \textit{corrupted} negative triple, for which we compute a score $s_n$. The loss is calculated as a function of $s_p$ and $s_n$. This approach is quite general and admits different loss and scoring functions. The complete procedure is presented in Algorithm \ref{alg:main}.

\begin{algorithm}[t]
\KwIn{Knowledge graph $(\mathcal{E},\mathcal{R},\mathcal{T},\mathcal{D})$, entity encoder $f_\theta$ with parameters $\theta$, learning rate $\eta$, scoring function $s$, loss function $\mathcal{L}$}
 $\mathbf{r}_j \gets$ initialize randomly for each $r_j\in\mathcal{R}$\;
 $\boldsymbol{\theta} := \lbrace \theta \rbrace \cup \lbrace\mathbf{r}_j \vert r \in \mathcal{R}\rbrace$\;
 \For{$(e_i, r_j, e_k)\in\mathcal{T}$}{
  $(e_i', r_j, e_k') \gets$ corrupt $(e_i, r_j, e_k)$\;
  $s_p \gets s(f_\theta(d_{ei}), \mathbf{r}_j, f_\theta(d_{ek}))$\;
  $s_n \gets s(f_\theta(d_{ei'}), \mathbf{r}_j, f_\theta(d_{ek'}))$\;
  $\boldsymbol{\theta}\gets \boldsymbol{\theta} - \eta\nabla_{\boldsymbol{\theta}}\mathcal{L}(s_p, s_n)$
 }
 \Return{$\boldsymbol{\theta}$}
 \caption{Learning inductive entity representations via link prediction}
 \label{alg:main}
\end{algorithm}

Note that our proposed algorithm is fundamentally different from KEPLER~\citep{wang2019kepler}, which is trained with an additional language modeling objective that is more expensive to compute, and requires more training data.

\subsection{Computational complexity}

A significant portion of the cost in Algorithm \ref{alg:main} comes from the entity encoder. Encoding a sequence of length $n$ with BERT has a complexity of $O(n^2)$, thus the time complexity for training is $O(\vert \mathcal{T}\vert n^2)$. In practice, a fixed value of $n$ can be selected (such as 32 or 64 in our experiments), so if we consider it equal for all entities, the algorithm remains linear with respect to the number of triples in the graph, up to a constant factor.

At test time, the embeddings for all entities can be pre-computed. In this case, link prediction for a given entity and relation is linear in the number of entities in the graph, and the entity encoder is only applied to new entities.

\subsection{Theoretical motivation}

Multiple models for modeling relations in KGs have been proposed as a form of factorization \citep{kolda2009tensor,nickel2015review}, by representing the relations in the graph as a third-order tensor $\mathbf{Y}\in\lbrace0, 1\rbrace^{\vert\mathcal{E}\vert\times\vert\mathcal{R}\vert\times\vert\mathcal{E}\vert}$, where the entry $y_{ijk}=1$ if $(e_i, r_j, e_k) \in\mathcal{T}$, and $y_{ijk}=0$ otherwise. For each $r_j\in\mathcal{R}$, $\mathbf{E}\mathbf{R}_j\mathbf{E}^\top$ is a \textit{tensor decomposition} of $\mathbf{Y}$, where $\mathbf{E}\in\mathbb{R}^{\vert\mathcal{E}\vert\times d}$ and $\mathbf{R}_j\in\mathbb{R}^{d\times d}$, and the $i$-th row of $\mathbf{E}$ contains the embedding $\mathbf{e}_i$ for $e_i\in\mathcal{E}$. Examples of models optimized for this kind of decomposition are RESCAL \citep{nickel2011three}, DistMult \citep{yang2015embedding}, ComplEx \citep{trouillon2016complex}, and SimplE \citep{kazemi2018simple}.

For an entity description $d_{ei} = (w_1, \dots, w_n)$, let $\mathbf{W}_{ei}\in\mathbb{R}^{d\times n}$ be a matrix of word embeddings, with the embedding of word $w_j$ in the $j$-th column. Approximating such a decomposition with an entity encoder thus requires correctly mapping $\mathbf{W}_{ei}$ to the embedding of the entity $\mathbf{e}_i$ in $\mathbf{E}$. In a recent result, \citet{Yun2020Are} show that Transformers are universal approximators of continuous functions with compact support\footnote{Their results are shown for functions whose range is $\mathbb{R}^{d\times n}$, but here we state a special case where we select one column from the output.} $g:\mathbb{R}^{d\times n}\rightarrow\mathbb{R}^{d}$. Therefore, if such a function exists so that $g(\mathbf{W}_{ei}) = \mathbf{e}_i$, there is a Transformer that can approximate the corresponding tensor decomposition. While the existence of this function is not obvious, it further motivates an empirical study on the use of BERT for entity embeddings in a KG.

\section{Evaluating entity representations}
\label{sec:experiments}

The use of an entity encoder to obtain entity representations is a more flexible approach that would be useful not just for link prediction, but for other tasks that could benefit from a vector representation that is a function of the textual description of an entity.

To better explore the potential of such an approach, we propose an evaluation framework that comprises inductive link prediction, inductive entity classification, and information retrieval for entity-oriented search. We present the results of the encoder proposed in \autoref{sec:inductive}, and compare with recently proposed methods for each task. Our implementation and all the datasets that we use are made publicly available\footnote{\url{https://github.com/dfdazac/blp}}.

\subsection{Link prediction}

\begin{table}[t]
\caption{Statistics of datasets used in the link prediction task.}
\label{tab:stats}
\centering
\begin{tabular}{lccc}
\toprule
          & WN18RR & FB15k-237  & Wikidata5M \\
\midrule
Relations & 11     & 237        & 822        \\
\midrule
          &        & Training                \\
                \cmidrule(lr){2-4}                        
Entities  & 32,755 & 11,633     & 4,579,609  \\
Triples   & 69,585 & 215,082    & 20,496,514 \\
\midrule
          &        & Validation              \\
                 \cmidrule(lr){2-4}            
Entities  & 4,094  & 1,454      & 7,374      \\
Triples   & 11,381 & 42,164     & 6,699      \\
\midrule
          &        & Test                    \\
              \cmidrule(lr){2-4}            
Entities  & 4,094  & 1,454      & 7,475      \\
Triples   & 12,087 & 52,870     & 6,894      \\
\bottomrule
\end{tabular}
\end{table}

Link prediction models can be evaluated via a ranking procedure \citep{bordes2011learning}, using a test set of triples $\mathcal{T}'$ disjoint from the set of training triples $\mathcal{T}$. For a test triple $(e_i, r_j, e_k)$, a prediction for the tail is evaluated by replacing $e_k$ with an entity $\hat{e}_k$ in a set of incorrect candidates $\hat{\mathcal{E}}$, and a score is computed as $s(\mathbf{e}_i, \mathbf{r}_j, \hat{\mathbf{e}}_k)$. Ideally, all the scores for incorrect candidates should be lower than the score of the correct triple. A prediction for the head is evaluated similarly, by replacing $e_i$.

In the transductive setting, the entities in a test triple are assumed to be in the set of training entities. Furthermore, the set of incorrect candidates is the same as the set of training entities. In the inductive setting, we instead consider a test graph $(\mathcal{E}', \mathcal{R}', \mathcal{T}', \mathcal{D}')$. The sets of triples $\mathcal{T}'$ and $\mathcal{T}$ are disjoint, and for relations, we always assume that $\mathcal{R}' \subseteq \mathcal{R}$. According to the way the set of incorrect candidates $\hat{\mathcal{E}}$ is determined, we define two inductive evaluation scenarios:

\paragraph{Dynamic evaluation} In a test triple, a new entity may appear at the head, tail, or both positions. The set of incorrect candidates $\hat{\mathcal{E}}$ is the union of training and test entities, $\mathcal{E}\cup\mathcal{E}'$. This represents a situation in which new entities are added to the KG, and is challenging as the set of incorrect candidates is larger at test time than at training.
\paragraph{Transfer evaluation} In a test triple, both entities at the head and tail position are new, and the set of incorrect candidates $\hat{\mathcal{E}}$ is $\mathcal{E}'$, where $\mathcal{E}'$ is disjoint from the training set of entities $\mathcal{E}$. This represents a setting where we want to perform link prediction within a subset of entities, that was not observed during training. Such a situation is of interest, for example, when transferring a trained model to a specific sub-domain of new entities.

We consider that both scenarios are highly relevant for tasks of link prediction in incomplete KGs, in contrast with previous works that only consider one or the other~\cite{xie2016RepresentationLO,wang2019kepler}.

\begin{table*}
\caption{Results of filtered metrics for link prediction involving entities not seen during training. We use WN18RR and FB15k-237 for dynamic evaluation, and Wikidata5M for the transfer evaluation (see text for more details). Results for KEPLER are reported by \citet{wang2019kepler}.}
\label{tab:results-inductive}
\centering
\begin{tabular}{lcccccccccccc}
\toprule
              &         \mc{4}{c}{WN18RR}                 &        \mc{4}{c}{FB15k-237}               &       \mc{4}{c}{Wikidata5M}               \\ 
                        \cmidrule(lr){2-5}                         \cmidrule(lr){6-9}                          \cmidrule(lr){10-13}                
Method        &    MRR   &     H@1  &     H@3  &     H@10 &    MRR   &     H@1  &     H@3  &     H@10 &    MRR   &     H@1  &     H@3  &     H@10 \\ 
\midrule                                                                                                                                          
GloVe-BOW     &    0.170 &    0.055 &    0.215 &    0.405 &    0.172 &    0.099 &    0.188 &    0.316 &    0.343 &    0.092 &    0.531 &    0.756 \\ 
BE-BOW        &    0.180 &    0.045 &    0.244 &    0.450 &    0.173 &    0.103 &    0.184 &    0.316 &    0.362 &    0.082 &    0.586 &    0.798 \\ 
GloVe-DKRL    &    0.115 &    0.031 &    0.141 &    0.282 &    0.112 &    0.062 &    0.111 &    0.211 &    0.282 &    0.077 &    0.403 &    0.660 \\ 
BE-DKRL       &    0.139 &    0.048 &    0.169 &    0.320 &    0.144 &    0.084 &    0.151 &    0.263 &    0.322 &    0.097 &    0.474 &    0.720 \\
KEPLER        &     --   &     --   &     --   &     --   &     --   &     --   &     --   &     --   &    0.402 &    0.222 &    0.514 &    0.730 \\
\midrule                                                                                                                                          
BLP-TransE    &\bf 0.285 &    0.135 &\bf 0.361 &\bf 0.580 &\bf 0.195 &\bf 0.113 &\bf 0.213 &\bf 0.363 &    0.478 &    0.241 &    0.660 &    0.871 \\ 
BLP-DistMult  &    0.248 &    0.135 &    0.288 &    0.481 &    0.146 &    0.076 &    0.156 &    0.286 &    0.472 &    0.242 &    0.646 &    0.869 \\ 
BLP-ComplEx   &    0.261 &\bf 0.156 &    0.297 &    0.472 &    0.148 &    0.081 &    0.154 &    0.283 &    0.489 &    0.262 &\bf 0.664 &\bf 0.877 \\ 
BLP-SimplE    &    0.239 &    0.144 &    0.265 &    0.435 &    0.144 &    0.077 &    0.152 &    0.274 &\bf 0.493 &\bf 0.289 &    0.639 &    0.866 \\
\bottomrule
\end{tabular}
\end{table*}

\subsubsection{Experiments}

\paragraph{Datasets} We make use of FB15k-237 \citep{toutanova2015observed} and WN18RR \citep{dettmers2018convolutional}, which are datasets widely used in the link prediction literature. To obtain entity descriptions we employ the datasets made available by \citet{yao2019kg}. FB15k-237 is a subset of Freebase, where most entities correspond to people, movies, awards, and sport teams. Descriptions were obtained from the introduction section of the Wikipedia page of each entity. In WN18RR each entity corresponds to a word sense, and descriptions are their definitions. Instead of using the conventional splits used for these datasets, we implement a \textit{dynamic evaluation} scenario. We select 10\% of entities and their associated triples to form a test graph, 10\% for validation, and the remaining 80\% for training. At test time, all entities are used as incorrect candidates. For these datasets we choose a maximum length of entity descriptions of 32 tokens. This value was chosen as using more tokens did not bring significant improvements, while increasing monotonically the time required for training (we include details for these results in \autoref{app:length}).

For the \textit{transfer evaluation}, we present results on Wikidata5M, with the splits provided by \citet{wang2019kepler}. The graph is a subset of Wikidata, containing 4.6 million entities, and descriptions from the introduction section of Wikipedia. To further experiment with the scalability of our approach, we increased the description length to 64 tokens. Dataset statistics are listed in Table \ref{tab:stats}.

\paragraph{Experimental setup} Following the definition in \autoref{sec:blpdef}, we implement an entity encoder using the BERT-base configuration from the Transformers library \citep{Wolf2019HuggingFacesTS}, followed by a linear layer with 128 output units. 

We study the performance of our method in combination with four relational models: TransE, DistMult, ComplEx, and SimplE. We aim to cover early translational and multiplicative models (i.e. TransE and DistMult) as well as more recent models that have been shown to result on state-of-the-art performance for link prediction \cite{ruffinelli2020you} (i.e. ComplEx and SimplE). We denote the resulting models as BERT for Link Prediction (BLP) followed by the employed relational model (e.g. BLP-TransE).

As a baseline we consider DKRL, proposed by \citet{xie2016RepresentationLO}. In our implementation of DKRL, we use GloVe embeddings \citep{pennington2014glove} with a dimension of 300 for the input, and an output dimension of 128. We also reproduce their Bag-Of-Words (BOW) baseline, in which an entity is encoded as the average of embeddings of words in the description. We denote these models as GloVe-DKRL and GloVe-BOW, respectively. Following recent works on the properties and applications of static embeddings from the input layer of BERT \citep{peters2018dissecting,jawahar2019bertlang}, we also consider variants of the baselines that use context-insensitive BERT Embeddings (BE). We denote these as BE-DKRL and BE-BOW.

For all models, we run grid search using FB15k-237, and we select the hyperparameters with the best performance on the validation set. We reuse these hyperparameters for training with WN18RR and Wikidata5M, as we found that they performed well on these datasets.

For BLP models, in the grid search we consider the following -- loss function: \{margin, negative log-likelihood\}, learning rate: \{1e-5, 2e-5, 5e-5\}, L2 regularization coefficient: \{0, 1e-2, 1e-3\}. We use the Adam optimizer with a learning rate decay schedule with warm-up for 20\% of the total number of iterations. We train for 40 epochs with a batch size of 64 with WN18RR and FB15k-237, and 5 epochs with a batch size of 1,024 with Wikidata5M.

For the BOW and DKRL baselines the values for grid search were the following -- learning rate: \{1e-5, 1e-4, 1e-3\}, L2 regularization coefficient: \{0, 1e-2, 1e-3\}. We use Adam with no learning rate schedule, and we train the models for 80 epochs with a batch size of 64 with WN18RR and FB15k-237, and 10 epochs with a batch size of 1,024 with Wikidata5M.

The number of negative samples is 64 in all experiments.

\paragraph{Metrics} Given the score for a correct triple, and the corresponding set of negative scores obtained by replacing the head of the triple by an incorrect candidate, we sort them to obtain a ranked list. Let $r_{th}$ be the position of the correct triple in the rank. The \textit{reciprocal rank} is then $1/r_{th}$. This procedure is repeated by replacing the tail, to obtain the reciprocal rank $1/r_{tt}$. The Mean Reciprocal Rank is the mean of these two values, averaged across all triples in the knowledge graph:
\begin{equation}
    \text{MRR} = \frac{1}{2\vert\mathcal{T}\vert} \sum_{t\in\mathcal{T}}
    \left(
    \frac{1}{r_{th}} + \frac{1}{r_{tt}} 
    \right)
\end{equation}

The Hits at 1 metric (H@1) is obtained by counting the number of times the correct triple appears at position 1, and averaging as for the MRR. The H@3 and H@10 are computed similarly, considering the first 3 and 10 positions, respectively.

When scoring candidates for a given triple, we consider the \textit{filtered} setting \citep{bordes2013translating} where for each triple we consider as incorrect candidates all entities in the set $\hat{\mathcal{E}}$ minus those that would result in a correct triple, according to the training, validation, and test sets.

For more specific technical details on datasets and training, we refer the reader to Appendices \ref{app:technical} and \ref{app:computing}.

\subsubsection{Results}

We report the Mean Reciprocal Rank (MRR), and Hits at 1, 3, and 10, on the test set in Table \ref{tab:results-inductive}. For reference, we also show results reported by \citet{wang2019kepler} for KEPLER. We observe that in both the dynamic evaluation (WN18RR and FB15k-237) and the transfer evaluation (Wikidata5M), BLP-TransE consistently outperforms all the baselines across all metrics.

We note that TransE results in higher link prediction performance with BLP, compared to alternatives like DistMult, ComplEx, and SimplE in WN18RR and FB15k-237. ComplEx and SimplE improve on performance in Wikidata5M, which contains around two orders of magnitude more triples for training. This suggests that more elaborate relational models might be less data efficient compared with TransE, when used with BERT for link prediction. We note that the gap in performance between BLP-TransE and baselines is larger in WN18RR than in FB15k-237. We hypothesize that the definitions of words in WN18RR can have subtle variations of syntax that a BERT encoder captures better, while entities in FB15k-237 can be more easily identified by keywords, so that ignoring syntax does not result in a large drop in performance.

Interestingly, we observe that in Wikidata5M, KEPLER results in lower performance despite using a joint training objective that combines language modeling and link prediction. While it has been suggested that such an objective leads to increased performance~\cite{wang2019kepler}, here we observe that this is not the case: all BLP variants outperform KEPLER when using only the link prediction objective, which also results in a reduced computational cost during training.

Despite our best efforts, we could not find a DKRL model that performed better than BOW models. This is surprising since unlike DKRL, BOW models do not take word order into account. Interestingly, for both BOW and DKRL, BE models yields consistently better results than models using GloVe embeddings, while BE models use 80\% less parameters due to the use of WordPiece embeddings.

This can be attributed to differences in the data used to pre-train the embeddings, but more importantly, to the size of the embeddings: the size of BERT and GloVe embeddings is 768 and 300, respectively, which translates into a larger number of parameters in BE models. While GloVe uses one embedding per word and a vocabulary size of 400,000, BE models use wordpiece embeddings and a vocabulary size of 30,000. This means that BE models manage to reduce the number of parameters 80\% in comparison with GloVe models, while resulting in better link prediction performance.

\begin{figure}[t]
    \centering
    \includegraphics[width=0.80\linewidth]{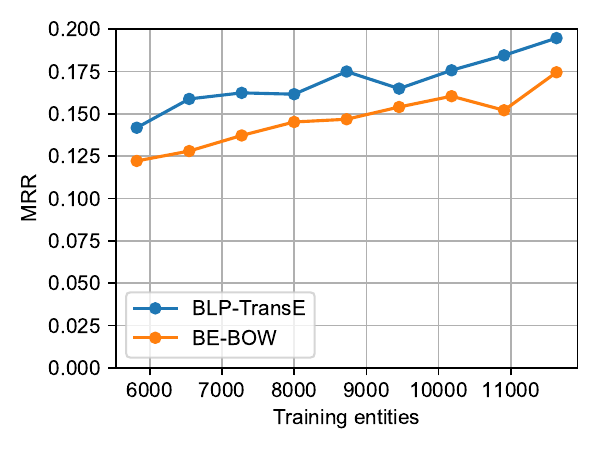}
    \caption{Inductive link prediction performance (MRR) versus number of entities used for training.}
    \label{fig:set-size}
\end{figure}

\paragraph{Effect of training set size} An important question in the inductive setting we consider is: what is the effect of the number of entities seen during training, on the performance on a fixed test set? To answer this question, we use FB15k-237 and the same test set, and sample subsets for training with an increasing number of entities. We evaluate the MRR for BE-BOW and BLP-TransE. The results are shown in \autoref{fig:set-size}. We note that a reduction of 50\% in the number of entities results in approximately a 27\% reduction in MRR. The mismatch between these percentages suggests that the pre-trained embeddings in BE-BOW and the architecture of BLP-TransE allow them to retain performance when reducing the size of the training set. \autoref{fig:set-size} also reveals in greater detail the constant gap between BE-BOW and BLP-TransE. Since both methods share the same pre-training regime and thus the data used during pre-training, we attribute the difference to the more powerful encoder used in BLP-TransE and the fact that it does not require dropping stopwords.

\paragraph{Transductive link prediction} Entity representations learned from entity descriptions can also be applied to the transductive scenario, where all entities at test time have been observed during training. This correspond to the setting where previous transductive methods for KG embeddings have been designed \cite{bordes2011learning,bordes2013translating,yang2015embedding,trouillon2016complex,kazemi2018simple}, although they cannot be applied to our experiments where textual descriptions and natural language are used. For reference, we include results in \autoref{app:transductive}, where we note that transductive methods significantly outperform description based encoders. We attribute this to the more challenging problem faced by description encoders: they must learn a complicated function from words to an entity representation, while transductive methods learn a lookup table with one embedding per entity and relation. Nonetheless, we stress that the applicability of description based encoders is much broader, as our work demonstrates.

\subsection{Entity classification}

A good description encoder must learn to extract the most informative features of an entity from its description, and compress them in the entity embedding. We test this property by using the embeddings of entities trained for link prediction, as features for a logistic regression classifier. Crucially, we maintain the inductive setting, keeping the splits from the link prediction experiments. Thus, at test time the classifier is evaluated on entities that the entity encoder did not see during training.

\subsubsection{Experiments}

\paragraph{Datasets} We evaluate entity classification using WN18RR and FB15k-237. In WN18RR we use the part of speech as the class for entities, which results in a total of 4 classes. For FB15k-237 we follow a procedure similar to \citet{xie2016RepresentationLO} by selecting the 50 most common entity types as classes.

\paragraph{Experimental setup} Using entity embeddings as features, we train a multi-class logistic regression classifier with L2 regularization. The regularization coefficient is chosen from \{1e-4, 1e-3, 1e-2, 0.1, 1, 10\}, and we keep the coefficient resulting in the best accuracy on the validation set. We also train classifiers with features not explicitly trained for link prediction: in GloVe-avg and BE-avg we use the average of GlovE and context-insensitive BERT embeddings, respectively. SBERT \citep{reimers2019sentence} is a model based on BERT that is trained to learn representations of sentences, that we apply to entity descriptions. We use their publicly available trained models\footnote{\url{https://github.com/UKPLab/sentence-transformers}} of the SBERT-NLI-base and SBERT-NLI-large variants.

\paragraph{Metrics} We report classification accuracy and its balanced version. The balanced accuracy weights each sample with the inverse prevalence of the true class, and allows us to identify when a classifier works better in average across classes, rather than performing better on the majority class.

\subsubsection{Results} We list the results in Table \ref{tab:results-classification}. We observe a drastic increase in performance with all BLP models trained for link prediction, which is especially noticeable when evaluating the balanced accuracy. The marked improvements in this metric demonstrate that the embeddings are a more informative representation that allows the classifier to perform better on classes for which there is very little data, and for entities not seen during training.

Interestingly, we note that i) the baselines not trained for link prediction perform better than the BOW and DKRL baselines in most cases, and ii) SBERT models still underperform BLP models trained for link prediction. We conclude that it is the combination of a powerful BERT encoder and a link prediction fine-tuning procedure that gives rise to better entity representations.

\begin{table}[t]
\caption{Accuracy for the entity classification experiments. Raw values correspond to the regular definition of accuracy. In the balanced case (Bal.), each sample is weighted with the inverse prevalence of its true class.}
\label{tab:results-classification}
\centering
\begin{tabular}{lccccc}
\toprule
                & \mc{2}{c}{WN18RR}   & \mc{2}{c}{FB15k-238}\\
Method          &    Raw   & Bal. &    Raw   & Bal. \\
\midrule
GloVe-avg       &    90.3 &    55.3 &    82.0 &    35.0 \\
BE-avg          &    92.7 &    62.1 &    82.4 &    39.4 \\
SBERT-NLI-base  &    96.3 &    66.5 &    84.5 &    36.6 \\
SBERT-NLI-large &    96.3 &    67.1 &    83.8 &    35.1 \\
\midrule                                                  
GloVe-BOW       &    91.5 &    56.0 &    82.9 &    34.4 \\ 
BE-BOW          &    93.3 &    60.7 &    83.1 &    28.3 \\ 
GloVe-DKRL      &    91.2 &    55.5 &    81.1 &    26.6 \\ 
BE-DKRL         &    90.0 &    48.8 &    81.6 &    30.9 \\ 
\midrule                                                  
BLP-TransE      &    99.1 &    81.5 &    85.4 &    42.5 \\ 
BLP-DistMult    &\bf 99.5 &    78.5 &    84.3 &    41.0 \\ 
BLP-ComplEx     &    99.3 &    78.1 &    85.1 &    38.1 \\ 
BLP-SimplE      &    99.2 &\bf 83.0 &\bf 85.8 &\bf 45.7 \\
\bottomrule
\end{tabular}
\end{table}

\subsection{Information retrieval}

An entity can be associated with different descriptions, that can be ambiguous and not necessarily grammatical. To evaluate the robustness of an entity encoder against this variability, we test its performance in an information retrieval task: given a query about an entity, return a list of documents (entity descriptions) ranked by relevance.

\begin{table*}
\caption{NDCG results for the information retrieval task, across different query types. We show the results for BM25F-CA, followed by the results after re-ranking with different entity encoders. Values in bold indicate that the difference between BM25F-CA and the re-ranked results is statistically significant at p $<$ 0.05.}
\label{tab:results-retrieval}
\centering
\begin{tabular}{lcccccccccc}
\toprule
                & \mc{2}{c}{SemSearch ES} & \mc{2}{c}{INEX-LD}    & \mc{2}{c}{ListSearch}  &  \mc{2}{c}{QALD-2}    &  \mc{2}{c}{All}       \\
Method          &      @10     &     @100 &    @10    &     @100  &     @10    &     @100  &    @10    &     @100  &    @10    &     @100  \\ 
\midrule                                                                                                                    
BM25F-CA        &      0.628   &    0.720 &    0.439  &    0.530  &     0.425  &    0.511  &    0.369  &    0.461  &    0.460  &    0.551  \\
\midrule
+ GloVe-BOW     &      0.631   &    0.721 &    0.449  &\bf 0.544  &     0.432  &\bf 0.518  &    0.368  &    0.460  &    0.462  &    0.554  \\
+ BE-BOW        &      0.629   &    0.721 &\bf 0.458  &\bf 0.546  &     0.431  &\bf 0.522  &    0.377  &    0.469  &    0.460  &    0.552  \\
+ GloVe-DKRL    &      0.624   &    0.719 &    0.440  &    0.529  &     0.424  &    0.516  &    0.368  &    0.468  &    0.459  &    0.550  \\
+ BE-DKRL       &      0.627   &    0.720 &    0.436  &    0.530  &     0.435  &\bf 0.525  &    0.374  &    0.466  &    0.459  &    0.553  \\
\midrule
+ BLP-TransE   &      0.631   &    0.723 &    0.446  &\bf 0.546  &     0.442  &\bf 0.540  &\bf 0.401  &\bf 0.482  &\bf 0.472  &\bf 0.562  \\
+ BLP-DistMult &      0.631   &    0.722 &\bf 0.458  &\bf 0.550  &     0.442  &\bf 0.536  &\bf 0.397  &\bf 0.480  &    0.468  &\bf 0.560  \\                                                                                         
+ BLP-ComplEx  &      0.628   &    0.721 &    0.454  &\bf 0.548  &     0.430  &\bf 0.528  &\bf 0.405  &\bf 0.486  &    0.468  &\bf 0.561  \\
+ BLP-SimplE   &      0.628   &    0.721 &    0.454  &\bf 0.552  &     0.439  &\bf 0.527  &\bf 0.399  &\bf 0.477  &    0.464  &\bf 0.557  \\
\bottomrule
\end{tabular}
\end{table*}

\subsubsection{Experiments}

\paragraph{Dataset} DBpedia-Entity v2 is a dataset for the evaluation of information retrieval (IR) systems for entity-oriented search, introduced by \citet{hasibi2017dbpediaent}. The document index corresponds to textual descriptions of entities in DBpedia. There are 467 queries, categorized into 4 types -- \textit{SemSearch ES}: short and ambiguous queries, e.g. ``john lennon, parents''; \textit{INEX-LD}: keyword queries, e.g. ``bicycle holiday nature''; \textit{List Search}: queries seeking for lists, e.g. ``Airports in Germany''; and \textit{QALD-2}: questions in natural language, e.g. ``What is the longest river?''. For each query, there is a list of documents graded by relevance by crowd workers. In average, there are 104 graded documents per query.

\paragraph{Experimental setup} Similarly as in previous work on embeddings for information retrieval \citep{gerritse2020graph}, we implement a re-ranking procedure, by updating a list of document scores assigned by an existing IR system (e.g. BM25). Let $q$ be query, $d_e$ the text in a document, and $z_{\text{IR}}$ the score the IR system assigned to $d_e$ given $q$. We use an entity encoder $f_\theta$ to compute the similarity between the embeddings of the query and the document, via their inner product:
\begin{equation}
z_{\text{new}} = \alpha f_\theta(q)^\top f_\theta(d_e) + (1 - \alpha) z_{\text{IR}}. 
\end{equation}
We select the best value of $\alpha$ via grid search on each of 5 training folds of the data provided by \citet{hasibi2017dbpediaent}, and report the average test fold performance. For the grid we consider 20 evenly spaced values in the interval $[0, 1]$, and for the entity encoder we use the models trained for link prediction with Wikidata5M. As in entity classification, we do not fine-tune the entity encoders. To obtain the base scores $z_\text{IR}$, we use BM25F-CA \citep{stephen2009bm25f}, as it is one of the best performing methods on the DBpedia-Entity v2 dataset reported by \citet{zhiltsov2015fsdm}.

\paragraph{Metrics} The DBpedia-Entity v2 dataset comes with a set of relevant documents for different queries. Each relevant document contains a grade of relevance, where 0 is non-relevant, 1 is relevant, and 2 is highly-relevant. This allows to compute how good the output ranking of a system is given the relevance of the documents it produces.

Assume we have a list of $k$ documents ranked by a system, and the relevance of the document at position $i$ is $r_i$. The Discounted Cumulative Gain (DCG) is defined as
\begin{equation}
    \text{DCG@k} = \sum_{i=1}^{k}\frac{2^{r_i} - 1}{\log_2(1 + i)}.
    \label{eq:dcg}
\end{equation}
Let IDCG@k be the maximum DCG@k, produced by sorting \textit{all} relevant documents in the corpus and computing eq. \ref{eq:dcg}. The Normalized DCG (NDCG) is defined as
\begin{equation}
    \text{NDCG@k} = \frac{\text{DCG@k}}{\text{IDCG@k}}.
\end{equation}

\subsubsection{Results} We report NDCG at 10 and 100 in Table \ref{tab:results-retrieval}. We show the results obtained with BM25F-CA, followed by the results after re-ranking with a particular encoder. BLP-TransE yields statistically significant improvements (according to a two-tailed t-test) in both NDCG at 10 and 100, when considering all query types.

Although on average the entity encoders yield higher performance across all query types, the difference with BM25F-CA is not statistically significant for SemSearch ES queries. This is expected since these queries are short and often not grammatical, differing from the entity descriptions used to train the encoders. In other query types the encoders show significant improvements, especially in QALD-2 queries, containing well formed questions.

Depending on the type of query, we observed that the optimal parameter $\alpha$ was always between 0.1 and 0.7, indicating that the embeddings of queries and documents alone are not enough to correctly rank documents, and a fraction of the score assigned by BM25F-CA is still required to preserve retrieval quality. However, the results of this section are encouraging in the context of representation learning of entities, as they show that the entity encoders have learned a function that maps entities and queries about them to vectors that are close in the embedding space.

We present a sample of an original ranking by BM25F-CA, and the reranked results by BLP-TransE in Table \ref{tab:query-demo}. In the first example, while BM25F-CA retrieves Intel-related companies and products, BLP-TransE fixes the ranking by pushing down products and increasing the scores for persons relevant to the query. In the second query, BM25F-CA retrieves more generic topics related to ``invention'', plus irrelevant documents at positions 2 and 5. BLP-TransE keeps only the first document and increases the scores of other, more relevant documents. 

\begin{table}
\caption{First five documents retrieved by BM25F-CA, and its reranking with BLP-TransE, for two example queries.}
\label{tab:query-demo}
\centering
\resizebox{0.48\textwidth}{!}{
\begin{tabular}{cll}
\toprule
\mc{3}{l}{\textbf{Query:} Who founded Intel?} \\
\midrule
Rank & BM25F-CA & + BLP-TransE \\
\midrule
1 & Intel & Intel \\
2 & Intel 8253 & Avram Miller \\
3 & Intel 8259 & Glenford Myers \\
4 & Intel Play & Intel Play \\
5 & Intel Ct   & Leslie L. Vadasz \\
\midrule
\mc{3}{l}{\textbf{Query:} A list of all American inventions} \\
\midrule
Rank & BM25F-CA & + BLP-TransE \\
\midrule
1 & US inventions & US inventions \\
2 & The Mothers of Invention &  US inventions (after 1991) \\
3 & American Heritage of Inv. & US inventions (1890 - 1945) \\
4 & Invention & Timeline of inventions \\
5 & Sandy Bull   & US inventions (1946-91) \\
\bottomrule
\end{tabular}
}
\end{table}

\section{Conclusion}

We have studied the generalization properties of entity representations learned via link prediction, by proposing the use of a pre-trained model fine-tuned with a link prediction, and an extensive evaluation framework that also considers tasks of entity classification and information retrieval. We find that when using the proposed model, the entity representations are inductive, as they exhibit strong generalization in link prediction for entities not seen during training.

Without requiring fine-tuning, the entity representations learned via link prediction are also transferable to the tasks of node classification and information retrieval, which demonstrates that the entity embeddings act as compressed representations of the most salient features of an entity. This provides evidence that the learned entity representations generalize to tasks beyond link prediction.

This is additionally important because having generalized vector representations of knowledge graphs is useful for using them within other tasks, for example, search and automatic metadata annotation. 

We consider the inductive setting where new entities appear at test time. Extending our approach to unseen relations is an interesting avenue for future research. For future work we also consider the study and improvement of the learned entity representations, for example, by enforcing independent factors of variation in the representations \citep{locatello2019disentangle}, or by learning on hyperbolic spaces better suited for hierarchical data \citep{nickel2017poincare}.

\section{Acknowledgements}

This project was funded by Elsevier’s Discovery Lab.

\bibliographystyle{ACM-Reference-Format}
\bibliography{sample-base}

\appendix

\section{Technical details}
\label{app:technical}

\paragraph{Dataset generation} The original splits of WN18RR and FB15k-237 are intended to be used in the transductive setting, where all entities in the validation and test sets are also present in the training set. We thus generate new splits for these datasets for the inductive setting. The algorithm for generating the datasets is as follows:
\begin{enumerate}
    \item Sample a node from the graph.
    \item Check that removing the node does not leave other nodes with zero neighbors.
    \item Check that if the node is removed, the number of edges for any relation type is not lower than 100.
    \item If conditions 2 and 3 are met, remove the node and add it to the set of nodes removed from the training set.
\end{enumerate}

\paragraph{Negative sampling} In all link prediction experiments, negative samples are obtained by randomly replacing the head or tail with a random entity. We sample entities from the set of entities involved in a mini-batch of triples. This strategy allows to increase the number of negative samples while preserving efficiency, as this requires encoding less distinct entity descriptions per mini-batch, than if the negative samples were sampled from the complete set of entities in the graph.

\section{Computing resources}
\label{app:computing}

For the experiments with WN18RR and FB15k-237 we used a workstation with an Intel Xeon processor, 1 NVIDIA GeForce GTX 1080 Ti GPU with 11GB of memory, and 60GB of RAM. For Wikidata5M we used 4 NVIDIA TITAN RTX GPUs with 24GB of memory each.

For all the BLP models we use the BERT-base configuration \citep{devlin2019bert}. In particular, we use the \texttt{bert-base-cased} pre-trained model from the Transformers library \citep{Wolf2019HuggingFacesTS}. This model has 110 million parameters. For link prediction, we additionally have parameters for relation embeddings, which have a dimension of 128. The number of parameters depends on the number of relations in the dataset. This results in 1,400 relation parameters for WN18RR, 30,300 for FB15k-237, and 105,216 for Wikidata5M.

The time required for training was 7 hours for WN18RR, 14 hours for FB15k-237, and 2 days for Wikidata5M.

\section{Effect of description length}
\label{app:length}

\begin{figure}[t]
    \centering
    \includegraphics[scale=0.7]{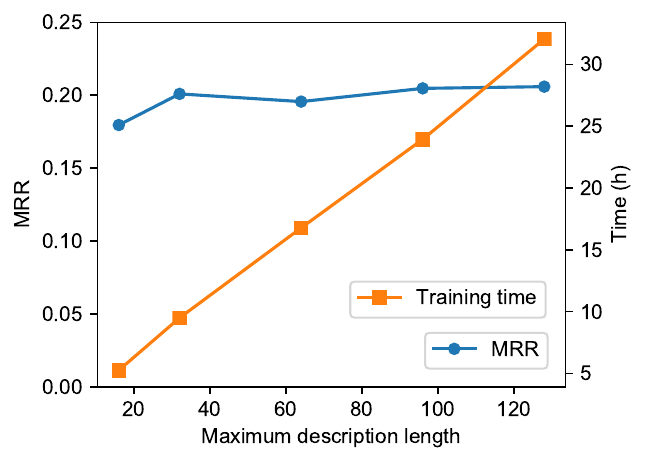}
    \caption{Inductive link prediction performance in the validation set and training time vs. maximum length of entity descriptions.}
    \label{fig:length}
\end{figure}

To determine the influence of the length of the descriptions on performance, we trained several variants of BLP-TransE with FB15k-237, varying the maximum length of the entity descriptions from 16 up to 128 tokens. The resulting link prediction performance (MRR) in the validation set, and training time are shown in \autoref{fig:length}. We note that using more than 32 tokens does not bring significant improvements, while monotonically increasing the time required for training.

\section{Transductive link prediction}
\label{app:transductive}

Entity representations learned via link prediction can also be applied to the transductive setting, where all entities at test time have been observed during training. For completeness, we compare the methods studied in our work, with previous works designed for this setting. The results are shown in \autoref{tab:results-transductive}.

\begin{table}[t]
\caption{Results of filtered metrics for link prediction in the transductive setting.  *Results taken from~\citet{sun2018rotate}. The first group contains transductive methods, and the second corresponds to inductive methods that use entity descriptions.}
\label{tab:results-transductive}
\centering
\resizebox{0.48\textwidth}{!}{
\begin{tabular}{lcccccccc}
\toprule
              &         \mc{4}{c}{WN18RR}                 &        \mc{4}{c}{FB15k-237}               \\ 
                        \cmidrule(lr){2-5}                         \cmidrule(lr){6-9}                
Method        &    MRR   &     H@1  &     H@3  &     H@10 &    MRR   &     H@1  &     H@3  &     H@10 \\ 
\midrule                                                                                              
TransE*       &    0.226 &      --  &      --  &    0.501 &    0.294 &      --  &      --  &    0.465 \\ 
DistMult*     &    0.430 &    0.390 &    0.440 &    0.490 &    0.241 &    0.155 &    0.263 &    0.419 \\ 
ComplEx*      &    0.440 &    0.410 &    0.460 &    0.510 &    0.247 &    0.158 &    0.275 &    0.428 \\ 
RotatE*       &\bf 0.476 &\bf 0.428 &\bf 0.492 &    0.571 &\bf 0.338 &\bf 0.241 &\bf 0.375 &\bf 0.533 \\
\midrule
KG-BERT~\cite{yao2019kg}   &    0.242 &    0.110 &    0.280 &    0.524 &    0.236 &    0.145 &    0.258 &    0.420 \\
BE-BOW        &    0.193 &    0.039 &    0.265 &    0.503 &    0.232 &    0.152 &    0.250 &    0.394 \\
BE-DKRL       &    0.222 &    0.047 &    0.323 &    0.558 &    0.215 &    0.135 &    0.231 &    0.379 \\
BERT-TransE   &    0.325 &    0.144 &    0.431 &\bf 0.679 &    0.235 &    0.150 &    0.253 &    0.411 \\ 
BERT-DistMult &    0.314 &    0.182 &    0.370 &    0.581 &    0.210 &    0.130 &    0.222 &    0.377 \\ 
\bottomrule
\end{tabular}}
\end{table}

\end{document}